\documentclass{article}
\usepackage[ruled,vlined]{algorithm2e}

\usepackage{subcaption}
\usepackage{graphicx}
\usepackage[compact]{titlesec}         

\AtBeginDocument{%
  \providecommand\BibTeX{{%
    \normalfont B\kern-0.5em{\scshape i\kern-0.25em b}\kern-0.8em\TeX}}}

\titlespacing{\section}{0pt}{0pt}{0pt} 
\AtBeginDocument{
  \setlength\abovedisplayskip{0pt}
  \setlength\belowdisplayskip{0pt}}
\begin{document}

\title{SepHRNet: Generating High-Resolution Crop Maps from Remote Sensing imagery using HRNet with Separable Convolution}
\author{Priyanka Goyal, Sohan Patnaik, Adway Mitra, Manjira Sinha}
\date{Indian Institute of Technology Kharagpur}
\maketitle

\begin{abstract}

The accurate mapping of crop production is crucial for ensuring food security, effective resource management, and sustainable agricultural practices. One way to achieve this is by analyzing high-resolution satellite imagery. Deep Learning has been successful in analyzing images, including remote sensing imagery. However, capturing intricate crop patterns is challenging due to their complexity and variability. In this paper, we propose a novel Deep learning approach that integrates HRNet with Separable Convolutional layers to capture spatial patterns and Self-attention to capture temporal patterns of the data. The HRNet model acts as a backbone and extracts high-resolution features from crop images. Spatially separable convolution in the shallow layers of the HRNet model captures intricate crop patterns more effectively while reducing the computational cost. The multi-head attention mechanism captures long-term temporal dependencies from the encoded vector representation of the images. Finally, a CNN decoder generates a crop map from the aggregated representation. Adaboost is used on top of this to further improve accuracy. The proposed algorithm achieves a high classification accuracy of 97.5\% and IoU of 55.2\% in generating crop maps. We evaluate the performance of our pipeline on the Zuericrop dataset and demonstrate that our results outperform state-of-the-art models such as U-Net++, ResNet50, VGG19, InceptionV3, DenseNet, and EfficientNet. This research showcases the potential of Deep Learning for Earth Observation Systems. 
\end{abstract}

\section{Introduction}

Spatiotemporal crop mapping is a significant area of research in remote sensing and agriculture that uses satellite imagery to identify and monitor the cultivation of crops over time and space. Accurate crop mapping is crucial for sustainable agriculture, as it helps optimize crop yields and increase food production. Additionally, crop mapping provides valuable insights into land-use changes, agricultural practices, and crop management. Furthermore, spatiotemporal crop mapping has applications beyond agriculture, including monitoring invasive species, urban growth, and changes in natural habitats, making it a valuable tool for environmental monitoring and management.

Recent advancements in crop mapping have been driven by the increasing availability of high-resolution satellite imagery and advances in machine learning algorithms. Deep learning models, such as Convolutional Neural Networks (CNNs), have improved the accuracy and efficiency of crop mapping from spatial satellite data by representing complex structures associated with different types of croplands.

Integration of multiple data sources, such as weather data, soil information, and topographic data, into crop mapping models, has further enhanced their accuracy and provided a more comprehensive understanding of crop growth and management, enabling more accurate predictions of crop yields and environmental impacts. There has also been a shift towards using spatio-temporal models that consider the dynamics of crop growth over time and space. These models provide insights into the effects of climate change, natural disasters, and land-use changes on crop production, informing strategies for adaptation.

In this study, we explore the task of spatio-temporal crop mapping from remote sensing images using several recent developments in Deep Learning, such as separable convolutions. We propose a pipeline that takes sequence of remote sensing images as input, incorporates High Resolution Network (HRNet) to capture spatial relations and an LSTM-based block and a self-attention mechanism to capture the temporal dependencies, to obtain a segmented image where each segment indicates a particular crop growth. Promising results are obtained on the publicly available ZueriCrop dataset, and several metrics are used to validate the robustness of the proposed pipeline.

The novelty of the proposed approach lies in our use of recent Deep Learning models and concepts for this task. We use the High Resolution Network (HRNet) and show its strong improvement in comparison to well-established image segmentation approaches such as U-Net. Further, we show that the use of separable convolution is far more effective for this task in comparison to traditional convolution. Further, we show that utilizing the sequential information is useful to create a more accurate representation of the crop map, and explore the use of sequential models like LSTM and Multi-Head Self-Attention.

The contributions of this work can be summarized as follows:
 \begin{enumerate}
\item We propose SepHRNet: an encoder-decoder based pipeline for generating high-resolution crop maps from remote sensing image sequence
\item We compare many recent Deep learning-based models at each step of the pipeline to choose the best one
\item We show that use of separable convolution instead of standard convolution and multi-head self-attention instead of LSTM improve the spatial and temporal representation respectively
\item We show that Boosting can help the models further
 \end{enumerate}

To establish the veracity of our contributions, we carry out extensive experiments on the ZueriCrop dataset, which contains sequences of remote sensing imagery over farmlands with ground-truth labels of crop production. We test different aspects of our proposed pipelines against alternate approaches. We consider and compare different deep learning architectures for the spatial component, as well as different convolution techniques. Regarding the temporal component, we compare LSTM and self-attention. Finally, we show how the use of Boosting (Adaboost) can further improve the performance of the proposed pipeline. We carry out a detailed ablation study to establish the importance of each part of the proposed pipeline.

The following section includes a description of prior work in the domain of mapping crop types using remote-sensing images in a spatiotemporal setting. Section 3 provides details about the dataset used along with data processing. In Section 4, the methodology is presented, including baselines and the proposed architecture, with a detailed mathematical representation. Training details, along with mainstream experiments, simulation results, and a performance comparison of the proposed model with existing ones, are explained in Section 5. Section 6 presents an ablation study conducted. Finally, the last section presents the conclusions drawn from this work.\\

\section{Related Work}
Crop mapping is an important task for agricultural planning. In recent years, remote sensing has emerged as a useful source of information for such crop mapping. Various techniques have been employed, including deep learning, time-series analysis, and machine learning, resulting in high classification accuracy for different crop types.

\subsection{Crop Mapping}
Mazzia et al. (2021) \cite{mazzia2020improvement} utilized multi-temporal Sentinel-2 imagery and crop information from the USDA National Agricultural Statistics Service (NASS) to train and evaluate their proposed spatiotemporal recurrent neural networks (STRNNs).
Turkoglu et al. (2021) \cite{turkoglu2021crop} introduced ms-convSTAR (multistage ConvRNN) and evaluated its performance on the Zuericrop dataset. They compared its performance with RF, LSTM, TCN, Transformer network, Unet, Unet + convLSTM, and Bi-convGRU.
Konduri et al. (2020) \cite{konduri2020mapping} employed the Cluster-then-label approach using Multivariate Spatio-Temporal Clustering and Mapcurves on the MODIS NDVI and USDA CDL dataset.
Rußwurm et al. (2019) \cite{russwurm2019breizhcrops} proposed the Breizhcrop time series dataset for crop identification and evaluated different models including RF, TCN, MSResNet, InceptionTime, OmniscaleCNN, LSTM, StarRNN, and Transformer.
Rustowicz et al. (2019) \cite{m2019semantic} introduced the first small-holder farms' crop type dataset of Ghana and South Sudan. They compared the performance of 2D U-Net + CLSTM, 3D CNN with RF, and bidirectional sequential encoder. 

Khaleque et al. (2020) \cite{khaleque2020crop} utilized Sentinel-2 time-series data and machine learning algorithms to classify crops, considering temporal variations. Chen et al. (2022) \cite{chen2022spatio} integrated convolutional neural networks (CNNs) and long short-term memory networks (LSTMs) for spatiotemporal crop mapping. 
Zhang et al. (2021) \cite{zhang2021spatiotemporal} combined spectral-temporal features and multi-scale spatial information using a multi-task CNN and morphological profile (MP) technique. Zhu et al. (2021) \cite{zhu2021spatio} employed a multi-scale CNN with random forest (RF) classification for crop mapping from multi-temporal Landsat imagery. 
Temporal variability of crop reflectance was considered by Liu et al. (2020) \cite{liu2020incorporating} using Sentinel-2 data and normalized difference vegetation index (NDVI) and enhanced vegetation index (EVI) as input features. Liu et al. (2021) \cite{liu2021temporal} studied the temporal consistency and variability of optical indices for crop mapping in Southwest China.
Yang et al. (2021) \cite{yang2021multi} proposed a multi-scale feature fusion approach for crop mapping using Sentinel-2 imagery. Hu et al. (2021) \cite{hu2021multi} used CNNs and a feature pyramid network (FPN) for multi-scale feature extraction. Shao et al. (2020) \cite{shao2020multi} combined spectral indices and image patches with a UNet architecture for crop classification.
These advancements are crucial for accurate crop mapping, enabling effective crop management and decision-making in agriculture.
\subsection{Deep Learning architectures}
VGG19 \cite{simonyan2014very} is a deep architecture with 19 layers, enabling it to learn hierarchical representations and capture complex patterns in crop images. However, its large number of parameters makes it computationally expensive, memory-intensive, and slower compared to other models. ResNet50 \cite{he2016deep} introduces residual connections that facilitate training deeper networks and capture discriminative features for crop mapping. However, its larger model size can be challenging in terms of memory usage and computational resources. InceptionV3 \cite{szegedy2015going} incorporates multi-scale feature extraction through inception modules with parallel convolutional layers of different sizes. This reduces the number of parameters, allowing for faster training and inference. However, multiple parallel convolutional layers increase computational complexity and may lead to information loss, although auxiliary classifiers help mitigate this issue. 

DenseNet121's \cite{huang2017densely} dense connectivity pattern allows for the direct flow of information between layers, enhancing gradient propagation and feature reuse. This improves parameter efficiency and captures fine-grained details and local features in crop images. However, direct connections increase memory usage during training and inference. EfficientNetV2 \cite{tan2019efficientnet} uses a compound scaling method to optimize resource allocation and achieve computational efficiency while maintaining accuracy. It incorporates Squeeze and Excitation (SE) blocks to capture important features and depthwise separable convolutions to reduce computational cost. However, the complex scaling coefficients may reduce model interpretability, and depthwise separable convolutions might impact the capture of complex relationships in crop images. HRNet \cite{sun2019deep} captures high-resolution details, multi-scale features, and contextual information. It maintains high-resolution representations throughout the network, captures fine-grained and coarse-grained features simultaneously, and integrates information from different levels of abstraction. However, it requires more computational resources, resulting in increased memory usage and longer training and inference times.

UNet \cite{ronneberger2015u} is effective in capturing fine details and spatial relationships within an image. The UNet architecture consists of an encoder and a decoder, with skip connections between corresponding layers in the encoder and decoder. The encoder captures hierarchical information at different scales, while the decoder upsamples the feature maps. The skip connections help preserve spatial information during upsampling. 
UNet++ \cite{zhou2018UNet++} builds upon the skip connections of the original UNet by introducing nested skip pathways, which allow for the integration of multi-scale contextual information. Each encoder block is connected not only to the corresponding decoder block but also to higher-resolution decoder blocks. This nested skip connection leverages multi-scale contextual information, enabling the network to capture both local and global contextual information more comprehensively. UNet++ offers an advanced and powerful architecture for crop mapping, allowing for more accurate and detailed segmentation of crops in satellite images or other remote sensing data.

Each of these deep learning models has unique architectural characteristics that make it suitable for crop mapping. These models can be trained to classify different types of crops or identify crop boundaries within an image. The input to the network is an image patch or a satellite image, and the output is a pixel-wise segmentation map where each pixel is assigned a class label representing the crop type or boundary. These models have demonstrated effectiveness in capturing spatial dependencies, contextual information, and fine-grained details, which are crucial for accurate crop mapping.\\

\section{Dataset: ZueriCrop}
\begin{figure}
\centering                                                     
\includegraphics[scale=0.75]{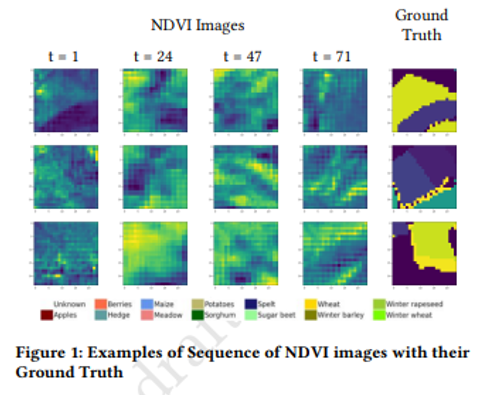}
\caption{Examples of Sequence of NDVI images (Columns 1-4) with their Ground Truth (Column 5)}
\label{fig:crop_lands}
\end{figure}

ZueriCrop \cite{turkoglu2021crop} is a large-scale, time-series dataset of satellite imagery of agricultural fields in Switzerland. It contains 116,000 field instances, each with ground-truth labels for 48 different crop types. The images were captured in 2019 under a variety of weather and lighting conditions, over a 50 km x 48 km area in the Swiss Cantons of Zurich and Thurgau. The dataset was made publicly available in 2021.
The images were acquired by the Sentinel-2 satellite, which provides high-resolution (10-meter) multispectral imagery. The images were atmospherically corrected using the standard Sen2Cor software package. 
Several crop land images can be seen in Figure \ref{fig:crop_lands}.

It is the largest publicly available dataset of time-series satellite imagery of agricultural fields. It contains a variety of crop types, some of which are difficult to distinguish from each other using satellite imagery. It may not be representative of agricultural practices in other parts of the world, since it was collected for a small area of Switzerland. Despite these challenges, ZueriCrop is a valuable resource for research in precision agriculture. 

Since the images are of $24X24$ resolution, and most of the deep learning architectures require a higher resolution of images, we resize the images using interpolation as well as padding. Moreover, the same transformation is also applied to the crop map in order to align the input to the output. After this, we normalize all the images by calculating the mean and the standard deviation of pixels per channel in order to make the pixel distribution uniform across channels.

\section{Methodology}
\label{sec:methodology}
The task of spatio-temporal crop mapping involves encoding the sequence of images, capturing temporal dependencies across the encodings, and finally obtaining the crop map as accurately as possible. We propose a deep learning-based solution to effectively capture temporal dependencies among the satellite images of land captured at different times of the year to obtain the crop distribution over that region. 
The rest of this section explains the proposed pipeline and its various parts that we explored to finally come up with a design that achieved the best performance on the ZueriCrop dataset.

\subsection{Pipeline Design}
\label{sec:baselines}
\begin{figure*}[htp]
\centering
\begin{subfigure}{0.3\textwidth}
    \includegraphics[width=0.9\textwidth]{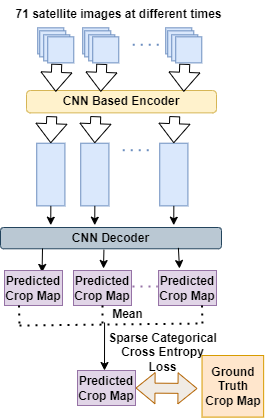}
    \subcaption{Encoder-Decoder Pipeline}
    \label{fig:encoder_decoder}
\end{subfigure}
\hspace{0.2cm}
\begin{subfigure}{0.3\textwidth}
    \includegraphics[width=1\textwidth]{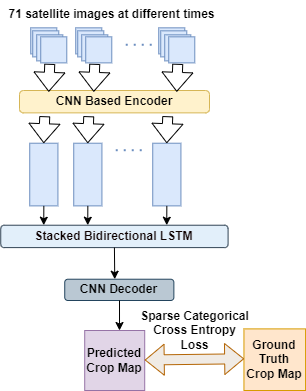}
    \subcaption{Encoder-LSTM-Decoder Pipeline}
    \label{fig:encoder_lstm_decoder}
\end{subfigure}
\hspace{0.4cm}
\begin{subfigure}{0.3\textwidth}
    \includegraphics[width=1\textwidth]{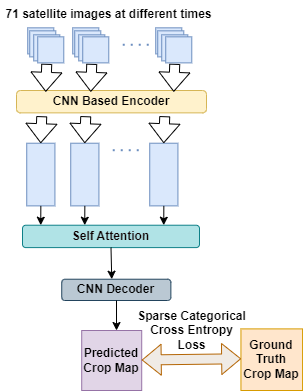}
    \subcaption{Encoder-SelfAttention-Decoder Pipeline}
    \label{fig:encoder_selfattention_decoder}
\end{subfigure}
\caption{Pipeline Design}
\label{fig:pipeline_design}
\end{figure*}

\textbf{Encoder - Decoder Architecture}
\label{sec:encoder_decoder}
The standard encoder-decoder pipeline can be used with the motivation to treat images at different time frames independently. This is basically the image segmentation problem of Computer Vision, for which there are well-known models that fit this approach are UNet \cite{ronneberger2015u}, and UNet++ \cite{zhou2018UNet++}. These can produce crop segmentation maps for the images at different time-points independently. Subsequently, we can compute the mean of all the resulting crop maps to aggregate the information and obtain the final crop map representing the land cover. The overview of this pipeline can be seen in Figure \ref{fig:encoder_decoder}.

\textbf{Encoder - LSTM - Decoder Architecture}
\label{sec:encoder_lstm_decoder}
An alternative paradigm of architecture is where we utilize the sequential relation between the images directly. Here, we can use convolutional neural networks such as VGG19 \cite{simonyan2014very}, ResNet50 \cite{he2016deep}, InceptionV3 \cite{szegedy2015going}, DenseNet121 \cite{huang2017densely}, EfficientNetV2 \cite{tan2019efficientnet}, or HRNet \cite{sun2019deep} to encode and obtain vector representation of the ZueriCrop images. VGG19, ResNet50, InceptionV3, DenseNet121, and EfficientNetV2 are pre-trained on ImageNet to learn generic features, that can be fine-tuned for crop mapping. They capture spatial dependencies and contextual information, which are crucial for accurate crop mapping. Next, a sequential model like LSTM \cite{hochreiter1997long} can be used in order to establish temporal relationships and obtain an aggregated representation of the land cover over the specified time frame. 
We found that a stacked LSTM layer with 3 blocks gives best results.
Finally, the last hidden state of the LSTM block is fed to a Transposed Convolution \cite{dumoulin2016guide} based decoder in order to obtain the segmented crop map. The overview of this architecture can be seen in Figure \ref{fig:encoder_lstm_decoder}.

\textbf{Encoder - Self Attention - Decoder Paradigm}
\label{sec:encoder_selfattention_decoder}
In this paradigm of architecture, we can use the same encoder and decoder types proposed in Section \ref{sec:encoder_lstm_decoder}. However, instead of using a stacked LSTM block for temporal modeling, we use Self Attention \cite{vaswani2017attention} mechanism to aggregate the vector representation of the images from different time-points. The architecture overview of this pipeline can be seen in Figure \ref{fig:encoder_selfattention_decoder}.

\subsection{Proposed Pipeline Design}
\label{sec:proposed_architecture}
\begin{figure*}[htp]
\centering
\begin{subfigure}{0.45\textwidth}
    \includegraphics[width=0.6\textwidth]{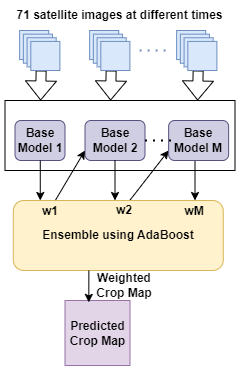}
    \subcaption{Overall}
    \label{fig:overall}
\end{subfigure}
\begin{subfigure}{0.45\textwidth}
    \includegraphics[width=1\textwidth]{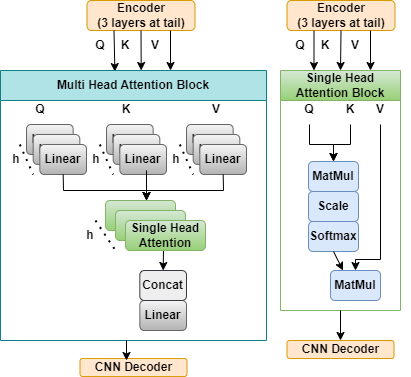}
    \subcaption{Self Attention}
    \label{fig:attention}
\end{subfigure}
\vspace{0.1cm}  
\begin{subfigure}{0.8\textwidth}
    \includegraphics[width=1\textwidth]{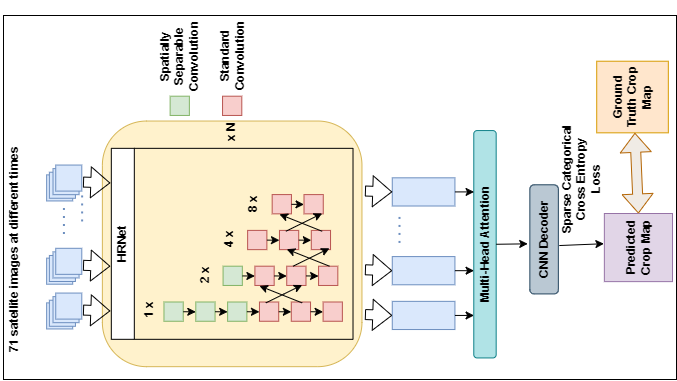}
    \subcaption{Base Model}
    \label{fig:base_model}
\end{subfigure}
\vspace{0.1cm}  
\begin{subfigure}{1\textwidth}
    \includegraphics[width=1\textwidth]{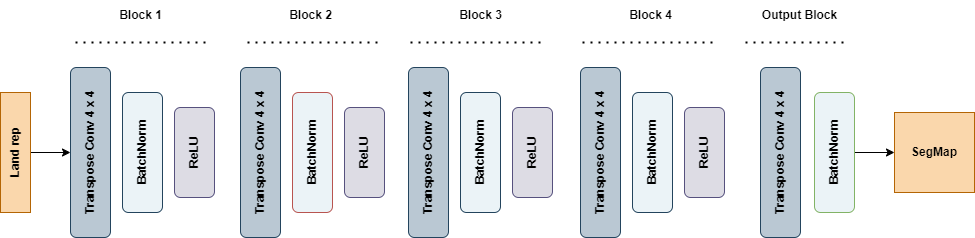}
    \subcaption{CNN Decoder}
    \label{fig:decoder}
\end{subfigure}    
\caption{Proposed Architecture}
\label{fig:proposed_architecture}
\end{figure*}

After conducting exhaustive experimentation and hyperparameter tuning, we have developed a pipeline that achieves the best performance among all the candidates previously discussed. The overview of our proposed pipeline is illustrated in Figure \ref{fig:proposed_architecture}.

We choose the Encoder-Self Attention-Decoder pipeline. In the encoder, we employ HRNet \cite{sun2019deep} to create a high-resolution representation of the satellite images at each time-point.
In the shallow layers of the network, we utilize Spatially separable convolution to reduce the model's parameter count while maintaining performance. By obtaining vector representations of all image frames, we leverage multi-head attention \cite{vaswani2017attention} to capture long-term temporal dependencies and generate a comprehensive representation of the land over the entire time-period. The aggregated representation is then fed into the decoder to produce the crop map. We call this combined model as \textbf{SepHRNet}.

We further improve performance using Boosting. SepHRNet serves as the base model for the AdaBoost \cite{freund1997decision} algorithm, with a modified rule for updating the sampling probability of data points. 
We combine multiple versions of SepHRNet trained on 80\% of the data, using the weighted ensemble as specified by the AdaBoost algorithm. Further, the Loss Function for AdaBoost is designed as follows: each image are assigned an error value of $1$ if the percentage of misclassified pixels is more than 20\%, and $-1$ otherwise. As a result, the aggregated model demonstrates strong performance across the entire dataset. 

\subsection{Components of Proposed Architecture}

\textbf{HRNet Encoder}
Let x be the input to the HRNet network.
\begin{equation}
HRNet(x) = H_n(H_{n-1}(...H_2(H_1(x))...))
\end{equation}
where $H_i$ denotes the $i_{th}$ stage of the HRNet network. Each stage consists of parallel branches, denoted as $B_i$, which operate on different resolutions of the input feature maps. The outputs of the branches in each stage are then combined to obtain the output of that stage. The HRNet network iteratively applies the stages $H_i$ to the input x, with the final output being the output of the last stage $H_n$. This allows the network to capture and integrate features at multiple resolutions, enabling it to maintain high-resolution representations throughout the network.

\textbf{Spatially separable Convolution}
\label{sec:separableConvolution}
Convolution is a well-known technique in image processing, which is widely used in Convolutional Neural Networks for image representation. Here we have a rectangular kernel $w$, which represents a spatial pattern, and we scan the image with it to see which parts of it contains that pattern. 
\begin{equation}
 y(i, j) = \sum_{m}\sum_{n} x(i-m, j-n) \cdot w(m, n) 
\end{equation}
A typical CNN has many layers for convolution, each of which uses many kernels for parallel channels. The parameters $w$ are not specified but learnt from data while training the neural network.  

Spatially separable convolution is a convolutional technique that offers advantages over standard convolution, particularly in scenarios with high aspect ratio images or when applying filters to small image regions. By using different kernel sizes for the vertical and horizontal dimensions, it can reduce the number of parameters in a convolutional neural network (CNN) and improve generalization performance by avoiding overfitting.
\begin{equation}\label{eq:sepcon}
z(i, j) = \sum_{m, n} x(i-m, j-n) \cdot w_{row}(m) \cdot w_{col}(n) 
\end{equation}
Equation~\ref{eq:sepcon} represents the spatially separable convolution operation where $z$ is the output obtained by convolving the input $x$ with the row-wise filter $w_{row}$ and the column-wise filter $w_{col}$. The summation is performed over the filter dimensions $m$ and $n$, and the element-wise multiplication of the input and filters is performed at each spatial location $(i, j)$.

In SepHRNet, we replace the standard convolution in shallow layers with spatially separable convolution. This replacement involves using two sequential convolutional layers with kernel sizes of $kX1$ and $1Xk$, respectively, instead of a single $kXk$ kernel. 
This modification maintains the same receptive field, reduces parameter count, and promotes more comprehensive interactions among pixels. As a result, our segmentation performance improves, especially considering the non-uniform distribution of land cover.

\textbf{Self-Attention}
Let $q_t \in \mathcal{R^{d_k}}$, $k_t \in \mathcal{R^{d_k}}$, and $v_t \in \mathcal{R^{d_k}}$ represent the query, key, and value vectors, respectively, at time step $t$. Matrix representations of the query, key, and value vectors are denoted as $Q = [q_1, q_2, \dots, q_T]$, $K = [k_1, k_2, \dots, k_T]$, and $V = [v_1, v_2, \dots, v_T]$, respectively. To compute the attention-weighted representation at a specific time step, we use the following equation:
        
\begin{equation}
Attention(Q, K, V) = softmax\left(\frac{QK^T}{\sqrt{d_k}}\right)V
\end{equation}

Here, $d_k$ represents the dimension of the query, value, and key vectors, i.e., the dimension of the output vector obtained from the encoder. 
The Attention function performs scaled dot-product attention, where the queries and keys are scaled by the square root of the key dimension $(d_k)$ and the result is weighted by the softmax of the query-key dot product. The final output is obtained by multiplying the weighted values with the softmax weights.

In order to incorporate self-attention, we require three types of vectors: query, key, and value for each input in the sequence. To obtain these vectors, instead of a single layer at the end of the encoder, we utilize three feed-forward layers. This allows us to generate the necessary query-key-value triplet of vectors. 
By summing the attention-weighted vectors, we obtain the aggregated representation of the cropland.
Figure \ref{fig:attention} illustrates the single-head attention block.

\textbf{Multi-head Attention}
Instead of having just one query-key-value triplet from the encoder, we obtain multiple triplets and compute the aggregated representation of all the queries, keys, and values. These representations are then concatenated and projected to the required dimension, resulting in the multi-head representation of the cropland illustrated in Figure \ref{fig:attention}. This approach allows us to attach more importance to the more complex structures in the cropland. 
        \begin{equation}
        {head}_i = {Attention}(Q \cdot W_Q^i, K \cdot W_K^i, V \cdot W_V^i) 
        \end{equation}
        Each $head_i$ is computed by applying the Attention function to the transformed queries $(Q \cdot W_Q^i)$, keys $(K \cdot W_K^i)$, and values $(V \cdot W_V^i)$.
        \begin{equation}
        {MultiHead}(Q, K, V) = {FFN}({Concat}({head}_1, ..., {head}_h))
        \end{equation}
Here, ${head}_i$ represents the output of a single-head self-attention, and $FFN$ refers to a feed-forward network used to downsample the concatenated representation. The MultiHead function calculates the multi-head self-attention by concatenating the individual attention heads $(head_i)$ and applying a feed-forward network.

\textbf{Decoder}
The aim of the decoder is to create the segmentation map at the same resolution as the input images. We use transposed convolution for this purpose. The architecture details can be seen in Figure \ref{fig:decoder}.
\begin{equation}
X = ReLU(BN(ConvTranspose(Z, W, S, padding) + B))
\end{equation}

In the above representation, Z represents the input feature map, W denotes the learnable weights of the transposed convolution operation, S represents the stride, and padding refers to the amount of zero-padding applied to the input feature map. The ConvTranspose operation performs the transposed convolution operation on Z using W, S, and the specified padding. It upsamples the input feature map by performing the reverse of the convolution operation, effectively increasing its spatial dimensions. The resulting output feature map X is then obtained by adding a bias term B to the transposed convolution output. BN denotes the batch normalization block. ReLU is used for RELU actiVation Function.\\

\section{Experimental Evaluation of Pipeline}

\subsection{Training Procedure}
The proposed architecture is trained using the sparse categorical cross-entropy loss function, which compares the softmax probabilities ($P_p$) with the ground truth labels ($T_p$) for each class. The loss is calculated according to Equation (\ref{22}):

\begin{equation} \label{22}
L = -\sum_{p=1}^N(T_p * \log(P_p))
\end{equation}

In this equation, $P_p$ is obtained using the softmax function to normalize the class probabilities. The loss function assigns smaller values for smaller differences and larger values for larger differences between the predicted and actual values. The goal is to minimize the loss, with a perfect model achieving a loss value of zero.

The segmentation network is trained on a split of 80\% for the train set (20,982 inputs) and 20\% for the test set (6,995 inputs), with each input consisting of 71 images capturing different time frames.
During training, the network is optimized using the Adam optimizer with a learning rate of $1e-4$, a batch size of 128, and a weight decay of $0.0001$. These hyperparameters are determined through grid search cross-validation. A cosine learning rate scheduler is employed over 25 epochs, and the training utilizes two V100 32GB GPUs in a distributed setup.
Stacked bidirectional LSTM with three layers is explored to capture temporal dependencies. The LSTM has a hidden state and output size of 256, and the LSTM block excludes bias terms in linear activations.
The self-attention block employs the standard attention mechanism, producing an aggregated representation of 768-dimensional vectors, serving as the base version. The multi-head attention block utilizes six attention heads as the default configuration.
Five models are used as the default number of base models in the ensemble paradigm experiments.\\

\subsection{Comparative Analysis of Architectures}
Section \ref{sec:proposed_architecture} introduces the proposed approach, which achieves the best performance. Table \ref{tab:ESD+Sep+Adaboost} provides a summary of the experiments, utilizing multi-head attention with six attention heads. HRNet-base as the encoder outperforms other encoder variants. Moreover, incorporating the self-attention mechanism to capture temporal dependencies in the underlying data improves the performance of each individual model. The ensembling approach demonstrates strong performance across the entire dataset.
\begin{table*}[h]
    \centering
    \small
    \begin{tabular}{cc|ccccc}
        \hline
        \textbf{Encoder} & \textbf{Paradigm} & & & \textbf{Metrics}\\
        \textbf{(with version)}& & \textbf{Accuracy} & \textbf{Precision} & \textbf{Recall} & \textbf{F1-score} & \textbf{mIoU}\\
        \hline
        VGG19 & ESD & 0.769 & 0.42 & 0.431 & 0.425 & 0.322\\
        ResNet50 & ESD & 0.897 & 0.513 & 0.482 & 0.497 & 0.388\\
        InceptionV3 & ESD & 0.922 & 0.556 & 0.591 & 0.573 & 0.493\\
        DenseNet121 & ESD & 0.906 & 0.571 & 0.573 & 0.572 & 0.449\\
        EfficientNetV2 & ESD & 0.931 & 0.589 & 0.561 & 0.575 & 0.495\\
        HRNet-base & ESD & \textbf{0.975} & \textbf{0.701} & \textbf{0.733} & \textbf{0.717} & \textbf{0.552}\\
        \hline
    \end{tabular}
    \caption{Performance evaluation among various versions of proposed approach}
    \label{tab:ESD+Sep+Adaboost}
\end{table*}
The experimental analysis demonstrates the performance of different encoder architectures with the ESD paradigm for crop mapping tasks. Among the models, HRNet-base achieves the highest performance across all metrics, with an accuracy of 0.975, precision of 0.701, recall of 0.733, F1-score of 0.717, and mIoU of 0.552. The use of self-attention for sequence modeling instead of LSTM improves the performance of all models. This finding highlights the effectiveness of the ESD paradigm and the self-attention mechanism for accurate crop mapping, enabling precise crop segmentation and classification.
\begin{figure}[h]
    \centering
    \includegraphics[scale = 0.3]{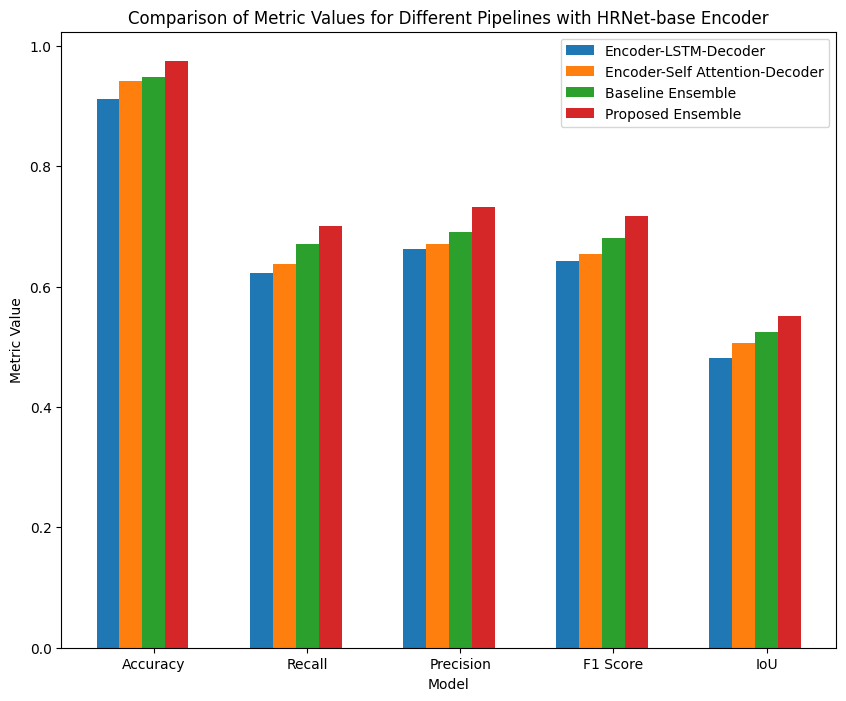}
    \caption{Comparision of metrics for HRNet-base encoder}
    \label{fig:HRNet_comparision}
\end{figure}
Figure \ref{fig:HRNet_comparision} provides a comparative overview of the improvement in different metrics when modifying the pipeline. It demonstrates that using self-attention instead of an LSTM block better captures temporal dependencies across time frames. Furthermore, combining spatially separable convolution and standard convolution in the encoder architecture enables the model to understand the underlying cropland with higher precision and accuracy. Based on the promising results of our proposed approach, we anticipate its generalizability to other datasets in the future.
\subsection{Visualization}
\begin{figure}
\centering
\begin{tabular}{@{}cccc@{}}\\
\textbf{\begin{tabular}[c]{@{}c@{}}Ground \\ Truth\end{tabular}}       & \includegraphics[scale=0.1]{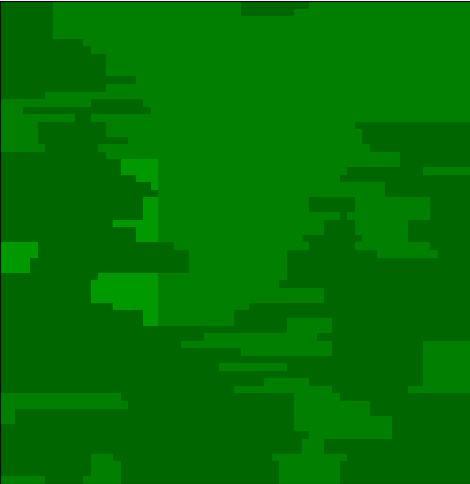} & \includegraphics[scale=0.1]{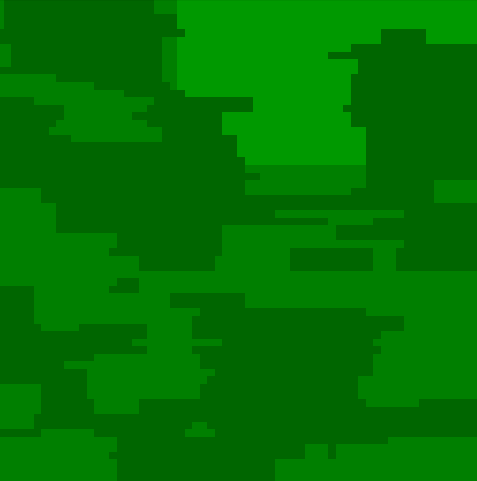} & \includegraphics[scale=0.1]{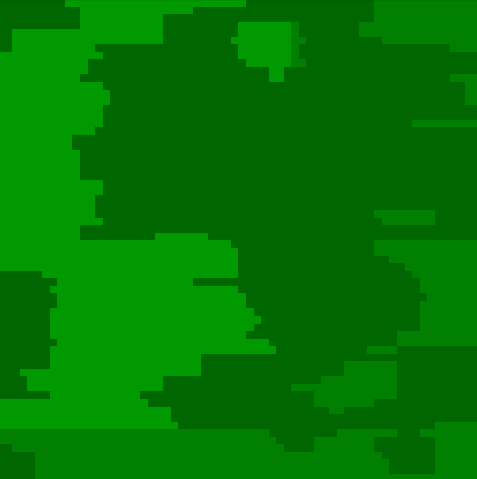}\\
\textbf{UNet} & \includegraphics[scale=0.1]{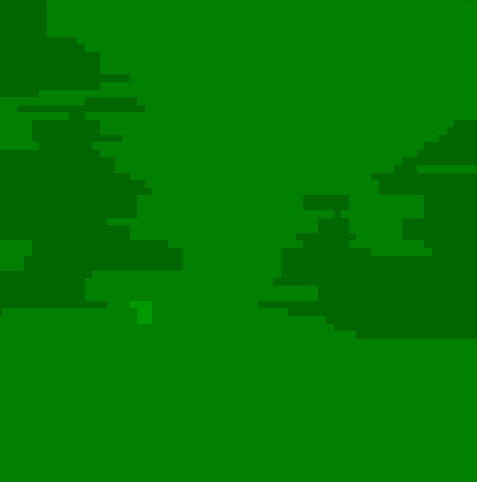} & \includegraphics[scale=0.1]{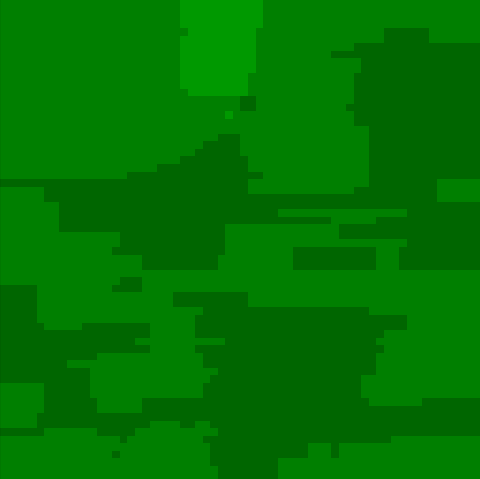} & \includegraphics[scale=0.1]{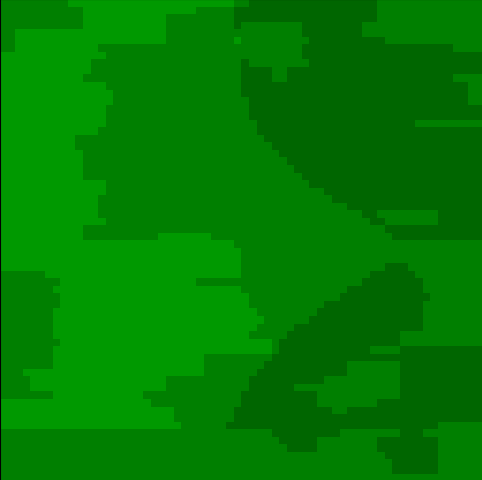} \\
\textbf{HRNet} & \includegraphics[scale=0.1]{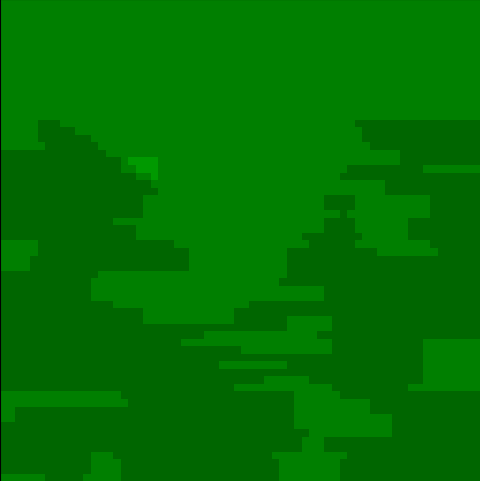} & \includegraphics[scale=0.1]{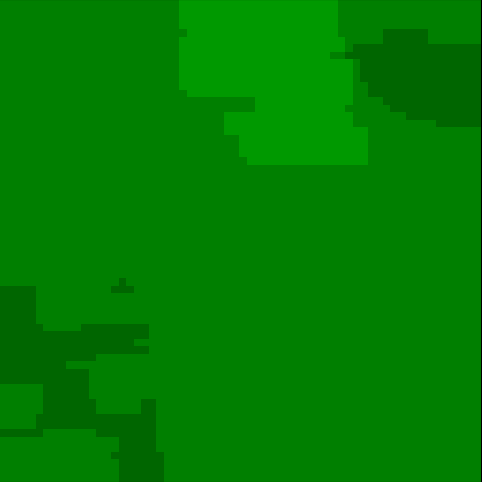} & \includegraphics[scale=0.1]{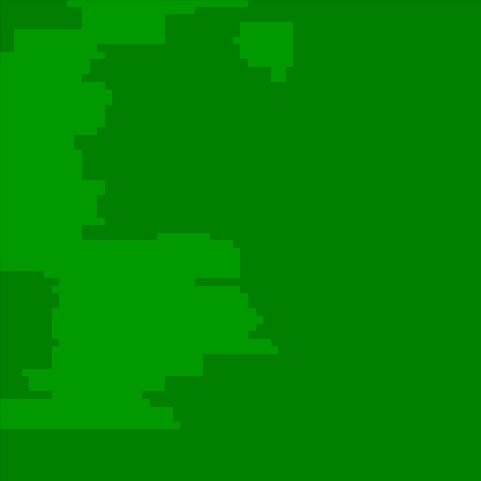} \\
\textbf{Proposed} & \includegraphics[scale=0.1]{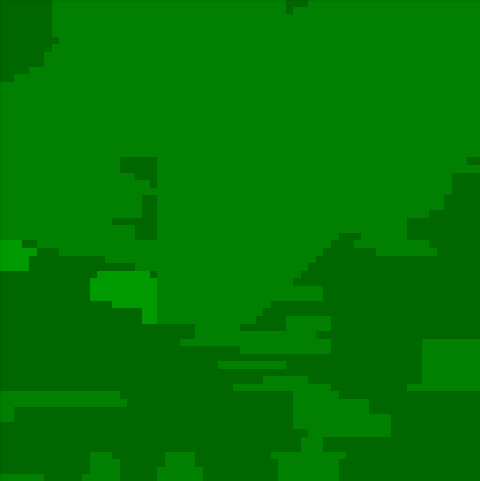} & \includegraphics[scale=0.1]{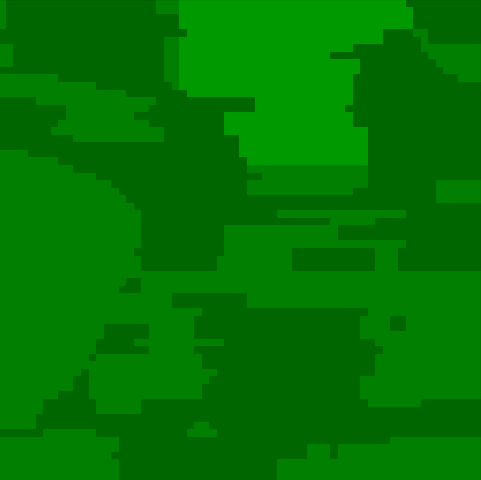} & \includegraphics[scale=0.1]{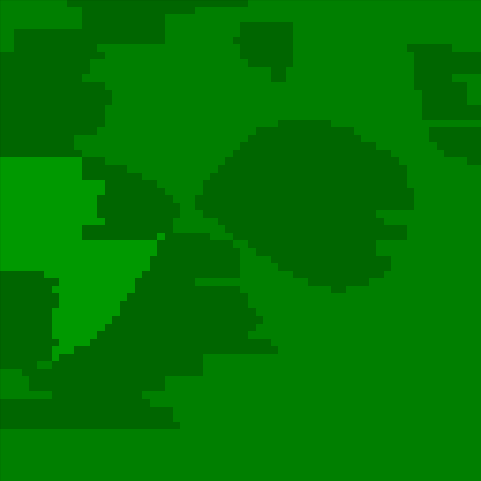} \\
\end{tabular}
\caption{ Predictions made on the test images of Zuericrop Dataset}
\label{fig:prediction}
\end{figure}

In this section, Figure \ref{fig:prediction} provides the visualization of some crop maps generated comparing the proposed model with baseline models HRNet and UNet. The proposed model seems to display better-quality crop maps on the ZueriCrop dataset. This serves as a motivation to incorporate Spatially separable convolution in place of the standard convolution of several standard encoder architectures. Moreover, ensembling the base models also shows promising results.

\section{Ablation Study} 
In this section, we present the results of various ablation experiments that demonstrate the enhanced performance of the models.

\subsection{Baselines}
In this section, we present the baseline results of the three proposed architectures discussed in Section \ref{sec:methodology}.
\begin{table*}[h]
    \centering
    \small
    \begin{tabular}{cc|ccccc}
        \hline
        \textbf{Encoder}  & \textbf{Paradigm} 	& & & \textbf{Metrics}\\
        \textbf{(with version)} & & \textbf{Accuracy} & \textbf{Precision} & \textbf{Recall} & \textbf{F1-score} & \textbf{mIoU}\\
        \hline
        VGG19 & ELD & 0.695 & 0.362 & 0.343 & 0.352 & 0.258\\
        ResNet50 & ELD & 0.831 & 0.464 & 0.455 & 0.459 & 0.331\\
        InceptionV3 & ELD & 0.852 & 0.473 & 0.484 & 0.478 & 0.352\\
        DenseNet121 & ELD & 0.834 & 0.461 & 0.469 & 0.465 & 0.337\\
        EfficientNetV2 & ELD & 0.851 & 0.476 & 0.492 & 0.484 & 0.363\\
        HRNet-base & ELD & 0.912 & 0.623 & 0.663 & 0.642 & 0.481\\  
        \hline
        UNet & ED & 0.846 & 0.473 & 0.468 & 0.470 & 0.352\\
        UNet++ & ED & 0.873 & 0.495 & 0.512 & 0.503 & 0.393\\
        \hline
    \end{tabular}
    \caption{Ablation 1: Baseline Paradigms}
    \label{tab:baseline_paradigm}
\end{table*}
In Table \ref{tab:baseline_paradigm},  "ELD", and "ED" refers to the Encoder-LSTM-Decoder, and Encoder-Decoder paradigm respectively. 
\textbf{Encoder - LSTM - Decoder Architecture}
\label{sec:enc_lstm_decoder}
To establish a baseline performance for spatio-temporal crop mapping and understand the behavior of different encoder architectures, we compare the results on the test set, as shown in Table \ref{tab:baseline_paradigm}. For all experiments, we use a stacked bidirectional LSTM with three layers.
Upon evaluating the performance of various encoder architectures, we find that the HRNet-base encoder significantly outperforms other versions. HRNet maintains multi-resolution inputs by fusing information from multiple resolutions in parallel, enabling the model to capture both fine-grained and coarse information in the image. It also enhances the localization of land patches, resulting in promising crop mapping results.

\textbf{Encoder - Decoder Architecture}
We enumerate the evaluation metrics for different encoder-decoder architectures in this subsection. As mentioned in Section \ref{sec:encoder_decoder}, we obtain the final crop distribution by taking the mean of the crop maps for each time frame. From Table \ref{tab:baseline_paradigm}, we observe that UNet++ outperforms UNet. Consequently, we choose HRNet-base as the preferred encoder due to its superior overall performance.

\subsection{Spatially Separable Convolution}
We evaluate the performance by replacing standard convolution with Spatially Separable Convolution in the encoder architecture. Table \ref{tab:axis_wise} illustrates  this change also improves the performance. By capturing features along one dimension before moving to the other, the model effectively captures the contours of the cropland, resulting in better outcomes. 
\begin{table*}[h]
    \centering
    \small
    \begin{tabular}{cc|ccccc}
        \hline
        \textbf{Encoder} & \textbf{Paradigm} & & & \textbf{Metrics}\\
        \textbf{(with version)}& & \textbf{Accuracy} & \textbf{Precision} & \textbf{Recall} & \textbf{F1-score} & \textbf{mIoU}\\
        \hline
        VGG19 & ELD & 0.721 & 0.388 & 0.401 & 0.394 & 0.297\\
        ResNet50 & ELD & 0.852 & 0.479 & 0.468 & 0.473 & 0.358\\
        InceptionV3 & ELD & 0.887 & 0.552 & 0.571 & 0.561 & 0.463\\
        DenseNet121 & ELD & 0.871 & 0.522 & 0.534 & 0.528 & 0.439\\
        EfficientNetV2 & ELD & 0.892 & 0.562 & 0.571 & 0.566 & 0.477\\
        HRNet-base & ELD & \textbf{0.941} & \textbf{0.662} & \textbf{0.691} & \textbf{0.676} & \textbf{0.524}\\
        \hline        
        UNet & ED & 0.876 & 0.542 & 0.577 & 0.559 & 0.479\\
        UNet++ & ED & 0.907 & 0.597 & 0.601 & 0.599 & 0.486\\
        \hline
    \end{tabular}
    \caption{Ablation 2: Incorporation of Spatially Separable Convolution}
    \label{tab:axis_wise}
\end{table*}
The experimental analysis shows that HRNet-base with the ELD paradigm achieves the highest performance in terms of accuracy, precision, recall, F1-score, and mIoU. It outperforms other encoder architectures in accurately classifying crop types and identifying crop boundaries. UNet++ with the ED paradigm also demonstrates competitive performance. The incorporation of spatially separable convolution improves the performance of both paradigms, highlighting its effectiveness in capturing fine-grained details and spatial relationships. Overall, HRNet-base with the ELD paradigm and UNet++ with the ED paradigm, incorporating spatially separable convolution, are effective for crop mapping tasks, providing accurate and detailed segmentation of crops in satellite images or other remote sensing data.

\subsection{Ensemble through Boosting}
We explore whether the ensemble strategy improves upon the baseline experiments by employing the AdaBoost algorithm. Table \ref{tab:ensemble} shows that the AdaBoost algorithm indeed enhances the performance of the corresponding baseline architectures. Among the ensembles, the HRNet-base ensemble demonstrates the best performance, as detailed in Section \ref{sec:enc_lstm_decoder}.
\begin{table*}[htp]
    \centering
    \small
    \begin{tabular}{cc|ccccc}
        \hline
        \textbf{Encoder} & \textbf{Paradigm} & & & \textbf{Metrics}\\
        \textbf{(with version)}& & \textbf{Accuracy} & \textbf{Precision} & \textbf{Recall} & \textbf{F1-score} & \textbf{mIoU}\\
        \hline
        VGG19 & ELD & 0.747 & 0.402 & 0.401 & 0.401 & 0.297\\
        ResNet50 & ELD & 0.863 & 0.482 & 0.468 & 0.475 & 0.358\\
        InceptionV3 & ELD & 0.895 & 0.531 & 0.571 & 0.550 & 0.463\\
        DenseNet121 & ELD & 0.892 & 0.552 & 0.534 & 0.543 & 0.439\\
        EfficientNetV2 & ELD & 0.921 & 0.582 & 0.571 & 0.576 & 0.477\\
        HRNet-base & ELD & \textbf{0.948} & \textbf{0.671} & \textbf{0.691} & \textbf{0.681} & \textbf{0.524}\\
        \hline
        UNet & ED & 0.881 & 0.561 & 0.577 & 0.569 & 0.479\\
        UNet++ & ED & 0.917 & 0.602 & 0.601 & 0.601 & 0.486\\
        \hline
    \end{tabular}
    \caption{Ablation 3: Ensembling}
    \label{tab:ensemble}
\end{table*}
The experimental analysis shows that HRNet-base with the ELD paradigm achieves the highest accuracy, precision, recall, F1-score, and mIoU among the encoder architectures and paradigms. The ensemble strategy improves the performance of all models, with HRNet-base and UNet++ demonstrating competitive results. This highlights the effectiveness of the ensemble strategy for crop mapping tasks. Overall, HRNet-base with the ELD paradigm and UNet++ with the ED paradigm, combined through an ensemble strategy, offer accurate and detailed crop segmentation, enabling precise crop classification and boundary delineation.

\subsection{Separable Convolution with Boosting}
We also see the impact of combining spatially separable convolution at the encoding stage, along with AdaBoost. We find that this results in improvement of performance in all baseline architectures of the encoder. 
\begin{table*}[htp]
    \centering
    \small
    \begin{tabular}{cc|ccccc}
        \hline
        \textbf{Encoder} & \textbf{Paradigm} & & & \textbf{Metrics}\\
        \textbf{(with version)}& & \textbf{Accuracy} & \textbf{Precision} & \textbf{Recall} & \textbf{F1-score} & \textbf{mIoU}\\
        \hline
        VGG19 & ELD & 0.752 & 0.41 & 0.4 & 0.405 & 0.299\\
        ResNet50 & ELD & 0.881 & 0.49 & 0.464 & 0.477 & 0.361\\
        InceptionV3 & ELD & 0.903 & 0.528 & 0.582 & 0.554 & 0.471\\
        DenseNet121 & ELD & 0.883 & 0.552 & 0.516 & 0.533 & 0.442\\
        EfficientNetV2 & ELD & 0.903 & 0.562 & 0.5558 & 0.560 & 0.464\\
        HRNet-base & ELD & \textbf{0.964} & \textbf{0.692} & \textbf{0.717} & \textbf{0.704} & \textbf{0.547}\\
        \hline        
        UNet & ED & 0.889 & 0.568 & 0.585 & 0.576 & 0.484\\
        UNet++ & ED & 0.92 & 0.613 & 0.605 & 0.609 & 0.493\\
        \hline
    \end{tabular}
    \caption{Ablation 4: Incorporation of Spatially separable convolution and ensembling approach}
    \label{tab:sep+ensemble}
\end{table*}
The experimental analysis for Table \ref{tab:sep+ensemble} reveals that HRNet-base with the ELD paradigm achieves the highest accuracy, precision, recall, F1-score, and mIoU among the different encoder architectures and paradigms. The incorporation of spatially separable convolution and the ensembling approach further improves the performance of all models. HRNet-base with the ELD paradigm achieves remarkable results with an accuracy of 0.964, precision of 0.692, recall of 0.717, F1-score of 0.704, and mIoU of 0.547. The UNet++ model with the ED paradigm also demonstrates competitive performance. This demonstrates the effectiveness of the spatially separable convolution and ensembling approach in enhancing crop mapping tasks. Overall, the combination of HRNet-base with the ELD paradigm and UNet++ with the ED paradigm, incorporating spatially separable convolution and employing an ensembling approach, provides accurate and detailed crop segmentation, enabling precise classification and boundary delineation of crops.

\subsection{Self-attention versus LSTM}
\begin{table*}[htp]
    \centering
    \small
    \begin{tabular}{cc|ccccc}
        \hline
        \textbf{Encoder}  & \textbf{Paradigm} 	& & & \textbf{Metrics}\\
        \textbf{(with version)} & & \textbf{Accuracy} & \textbf{Precision} & \textbf{Recall} & \textbf{F1-score} & \textbf{mIoU}\\
        \hline        
        VGG19 & ESD & 0.723 & 0.381 & 0.352 & 0.366 & 0.271\\
        ResNet50 & ESD & 0.857 & 0.472 & 0.648 & 0.546 & 0.353\\
        InceptionV3 & ESD & 0.871 & 0.513 & 0.509 & 0.511 & 0.394\\
        DenseNet121 & ESD & 0.842 & 0.476 & 0.482 & 0.479 & 0.361\\
        EfficientNetV2 & ESD & 0.875 & 0.494 & 0.511 & 0.502 & 0.403\\
        HRNet-base & ESD & \textbf{0.942} & \textbf{0.638} & \textbf{0.671} & \textbf{0.654} & \textbf{0.507}\\       
        \hline
    \end{tabular}
    \caption{Ablation 5: Using Self-attention for sequence modeling}
    \label{tab:baseline_paradigm_esd}
\end{table*}
Instead of using LSTM to capture temporal dependencies between cropland representations at different time frames, we employ a self-attention mechanism to better weigh the contribution of each representation. In all experiments listed in Table \ref{tab:baseline_paradigm_esd}, we utilize multi-head attention with six attention heads. As expected, HRNet-base as the encoder outperforms other encoder variants. Additionally, incorporating the self-attention mechanism improves the performance of each individual model, enhancing the capture of temporal dependencies in the data.

Its experimental analysis reveals that HRNet-base with the ESD paradigm achieves the highest accuracy, precision, recall, F1-score, and mIoU among the encoder architectures, making it an effective choice for crop mapping. Comparing the ELD and ESD paradigms, the ESD paradigm consistently outperforms the ELD paradigm across various encoder architectures in terms of accuracy, precision, recall, F1-score, and mIoU, indicating its effectiveness for crop mapping tasks. HRNet-base demonstrates superior performance in accurately classifying crop types and identifying crop boundaries compared to other encoders. UNet++ with ED, EfficientNetV2, and InceptionV3 with ESD also show competitive performance, while VGG19 exhibits lower performance. HRNet-base with the ESD paradigm emerges as a powerful choice, offering high accuracy, precision, recall, F1-score, and mIoU, which are crucial for precise crop classification and boundary delineation.\\

\section{Conclusion}
The aim of this work is to generate high-resolution crop maps based on remote sensing imagery. We use image sequences collected over a period of time, and aim to incorporate this temporal information into the model for more robust estimation of the segmented crop maps. For this purpose, we proposed a deep learning pipeline using Encoder-Self Attention-Decoder structure, which we named SepHRNet. For each of the parts, we compared multiple baselines on multiple criteria, and chose the best-performing options. In the encoder, HRNet along with spatially separable convolution is used, followed by Multi-Head Self-Attention followed by decoder based on Transposed Convolutions which produces the segmented map at the original resolution. Further, we see that the results can improve significantly by building an ensemble of SepHRNet by Adaboost. The pipeline was tested on the Zuericrop dataset for mapping 48 different types of crops over Switzerland. The proposed model demonstrated high accuracy, precision, recall, F1-score, and mIoU, making it an effective choice for crop mapping. This work highlights the importance of separable convolution for spatial modeling and multi-head self-attention for temporal modeling. Future work will include scaling up the proposed architecture for mapping of larger regions with more crop-types, and over different regions of the world. 

\bibliographystyle{ACM-Reference-Format}
\bibliography{Bibliography}

\end{document}